\documentclass[runningheads,a4paper]{llncs}

\usepackage{amssymb}
\usepackage{graphicx}

\usepackage{url}
\newcommand{\keywords}[1]{\par\addvspace\baselineskip
\noindent\keywordname\enspace\ignorespaces#1}

\begin{document}

\mainmatter  

\title{Lung Nodule Classification by the Combination of Fusion Classifier and Cascaded Convolutional Neural Networks}

\titlerunning{Multi-stage Neural Networks with Single-sided Classifiers}

%
%
\author{Masaharu Sakamoto\inst{1} \and Hiroki Nakano\inst{1} \and Kun Zhao\inst{2} \and Taro Sekiyama\inst{2}}

\authorrunning{Multi-stage Neural Networks with Single-sided Classifiers}

\institute{IBM Tokyo Laboratory, Watson Health, Tokyo, 103-8510, Japan \and IBM Research - Tokyo, Tokyo, 103-8510, Japan}

\maketitle

\begin{abstract}
Lung nodule classification is a class imbalanced problem, as nodules are found with much lower frequency than non-nodules. In the class imbalanced problem, conventional classifiers tend to be overwhelmed by the majority class and ignore the minority class. We showed that cascaded convolutional neural networks can classify the nodule candidates precisely for a class imbalanced nodule candidate data set in our previous study. In this paper, we propose Fusion classifier in conjunction with the cascaded convolutional neural network models. To fuse the models, nodule probabilities are calculated by using the convolutional neural network models at first. Then, Fusion classifier is trained and tested by the nodule probabilities. The proposed method achieved the sensitivity of 94.4\% and 95.9\% at 4 and 8 false positives per scan in Free Receiver Operating Characteristics (FROC) curve analysis, respectively.
\end{abstract}

\section{Introduction}
Lung nodule classification is a class imbalanced problem because nodules are found with much lower frequency than non-nodules. In the class imbalanced problem, conventional classifiers tend to be overwhelmed by the majority class and ignore the minority class. As one method to cope with the class imbalanced problem in lung nodule classification, we proposed cascaded convolutional neural networks that perform as single-sided classifiers for filtering out obvious non-nodules such as blood vessels or ribs \cite{sakamoto2017}. The method can achieve a few false positives while maintaining high sensitivity in the lung nodule classification. 
In this paper, we propose Fusion classifier that is a deep neural network to fuse the output of multiple models. To fuse models, nodule probabilities are calculated by using convolutional neural network models. Then, Fusion classifier is trained and tested by the nodule probabilities. Finally, an output of Fusion classifier is a nodule probability of the nodule candidate. The proposed method achieved a few false positives while maintaining higher sensitivity than using cascaded neural networks solely.

\section{PRIOR WORKS}
Several approaches have been proposed to deal with the class imbalanced problems in medical diagnosis \cite{rahman2013addressing}, detection of oil spills in satellite radar images \cite{kubat1998machine} and the detection of fraudulent calls \cite{fawcett1997adaptive}. Japkowicz \cite{japkowicz2000learning} showed that oversampling minority class and subsampling majority class are both very effective methods of coping with the problem. Chawla et al. \cite{chawla2002smote} proposed SMOTE (Synthetic Minority Over-Sampling Technique) algorithm that artificially creates minor class samples and randomly sub-samples majority class. Sun et al \cite{sun2009classification} comprehensively reviewed the class imbalanced problems. Kubat and Matwin \cite{Kubat97addressingthe} proposed a one-sided selection method that retains all minor class samples and sub-samples majority class. This study has similar object as us, but our method is completely different from the one-sided selection.   
In recent years, Convolutional Neural Network (CNN) has become available thanks to high speed and large capacity computing resources and it is showing superior performance to conventional technology in computer vision applications \cite{krizhevsky2012imagenet}. This is because CNN can be trained end-to-end in a supervised fashion while learning highly discriminative features, thus removing the need for handcrafting nodule feature descriptors. Viola and Jones \cite{viola2001rapid} and Wu et al. \cite{wu2003learning} used weak classifiers to construct boosted cascade layer with simple features. Compared with the weak classifiers with simple features, our method uses CNNs which are not weak classifiers and automatically capture features from input. As for the cascaded CNN structure, Li et al. \cite{li2015convolutional} proposed a cascaded CNN structure for face detection. They use 6 CNNs in the cascade including 3 CNNs to detect the face and 3 CNNs to calibrate the bounding box separately. The bounding box calibration is not needed in our proposed method. The applications of cascaded CNN for face detection \cite{qin2016joint}\cite{kalinovskii2015compact} and other kind of image feature detection \cite{chen2016mitosis} can also be found. However, the class imbalanced problem is not addressed in those works. 
Setio, et al. \cite{setio2016pulmonary} used a multi-stream CNN to classify points of interest in lung CT as a nodule or non-nodule. Dou et al. \cite{dou2016multi}  proposed a method employing 3D CNNs for false positive reduction in automated pulmonary nodule detection from volumetric CT scans. However, the class imbalanced problem is not addressed in those works

\section{CASCADED NEURAL NETWORKS}
\begin{figure}
\centering
\includegraphics[width=\textwidth]{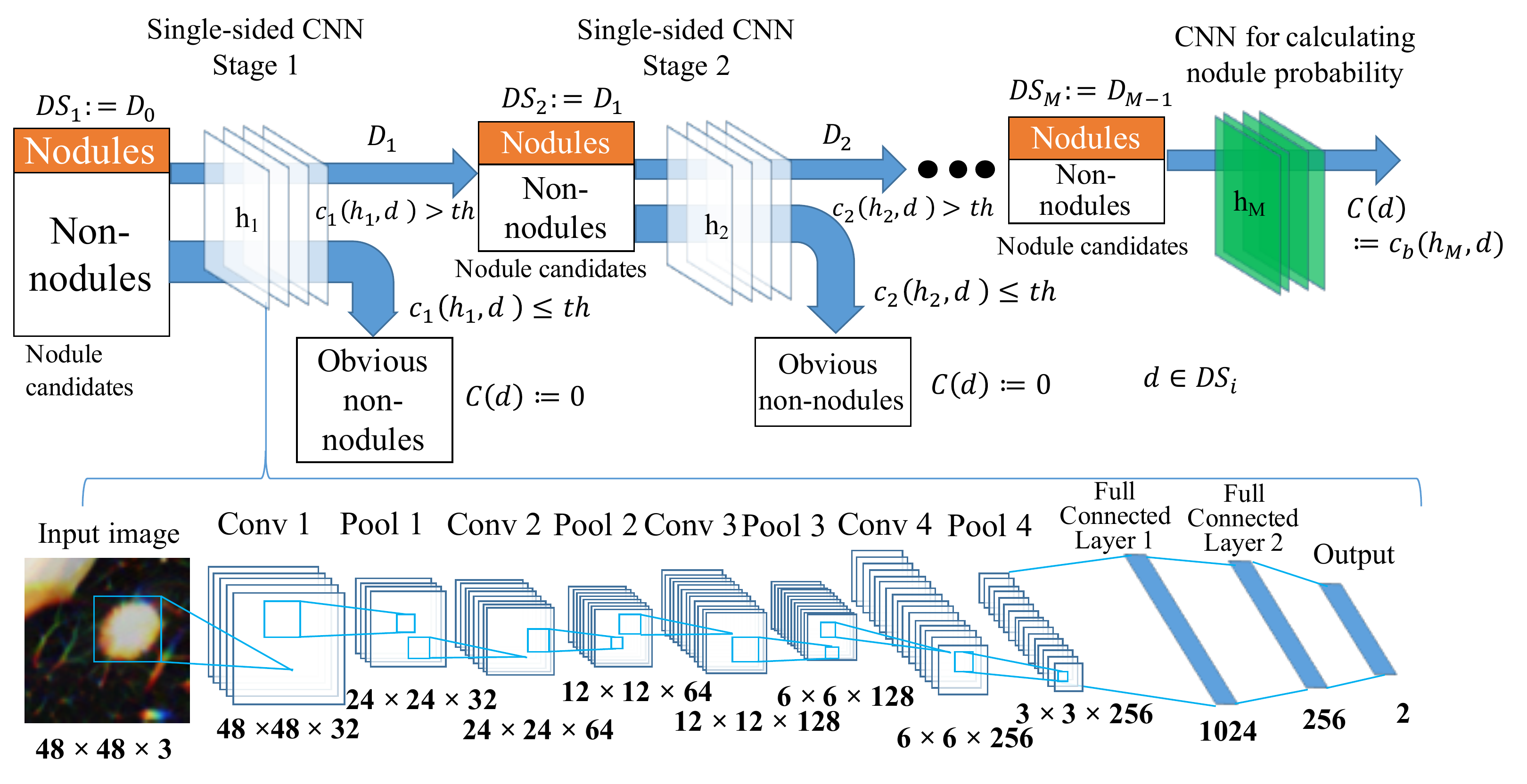}
\caption{Schematic diagram of cascaded multi-stage CNNs. Stage 1, Stage 2 and Stage n are CNNs that perform as single-sided classifiers to filter out non-nodule lesions. The final stage is a CNN to calculate nodule probabilities. $c(x)$ is nodule probability of nodule candidate $x$. $th$ is a threshold value to filter out obvious non-nodules. The lower part shows the structure of the CNN. The numbers at lowest part show number of neurons in three dimensions (width, height and channel).}
\label{fig:ccnn}
\end{figure}

We proposed cascaded neural networks to cope with the class imbalanced problem in lung nodule classification \cite{sakamoto2017}. The method consists of CNNs perform as single-sided classifiers, filtering processes for obvious non-nodules and a balanced CNN. To implement the single-sided classifiers, CNNs are trained with an inversely imbalanced data set consisting of many nodule images and a few non-nodule images. By “inverse” we mean that the ratio of the number of nodules and non-nodules is inversed against the original dataset. As the results, the single-sided classifiers work well for nodule samples, but do not work well for non-nodule samples. By using a thresholding in nodule probability, non-nodule samples are classified into obvious non-nodules and suspicious nodule candidates. Our method positively utilizes deterioration of classification performance caused by class imbalance learning. We call such classifier as single-sided classifier. The filtering process retains all nodules samples and sub-samples non-nodule samples.
Figure \ref{fig:ccnn} shows a schematic diagram of test method. Stage 1, stage 2 and stage $(M-1)$ are CNNs that perform as single-sided classifier. At stage $i \{i=1,2,\cdots,M-1\}$, for a nodule candidate $d \in DS_i$, nodule probability $c_i(h_i, d)$ is calculated by using a single-sided classifier model hi. Where, DSi is a test data set at stage i. A test sample whose probability falls below a threshold value th is filtered out and assigned zero probability value, $C(d) := 0$, as an obvious non-nodule. Test samples whose probability value exceeds the threshold value th are classified as suspicious nodule candidates Di, then handed over to the next stage. At the next stage $(i+1)$, the same procedures are applied to test data set $DS_i+1:=D_i$ again. The procedures are repeated until stage $(M-1)$. At stage $M$, a CNN model $h_M$, trained by a balanced data set, calculates nodule probability $c_b(h_M, d)$ for a suspicious nodule candidate d∈DSM that is handed over from the stage $(M-1)$. The $c_b(h_M, d)$ is assigned as a nodule probability of the sample d as $C(d):= c_b(h_M, d)$.
Training procedure is similar to the test procedure. An inversely imbalanced data set is extracted from a suspicious data set whose handed over from the previous stage, then used for training a single-sided classifier. The remaining data is used for test and to filter out obvious non-nodules by using the single-sided classifier. The test samples which were not filtered out are handed over to the next stage. Such procedures are repeated until stage $(M-1)$. At stage M, a balanced data set is extracted from the suspicious data set whose handed over from stage $(M-1)$, then used for training a CNN. We used a cross validation manner to implement the training procedure. 
As the result, CNN models which trained with different data set are created, and they have different properties. To take of advantage of the diversified models, we propose Fusion classifier. 

\section{Fusion classifier}
Fusion classifier is a deep neural network. Input of Fusion classifier is a nodule probability vector of a nodule candidate whose generated from multiple CNN nodule classifiers. Output of Fusion classifier is a nodule probability of the nodule candidate. 

\begin{figure}
\centering
\includegraphics[width=0.6\textwidth]{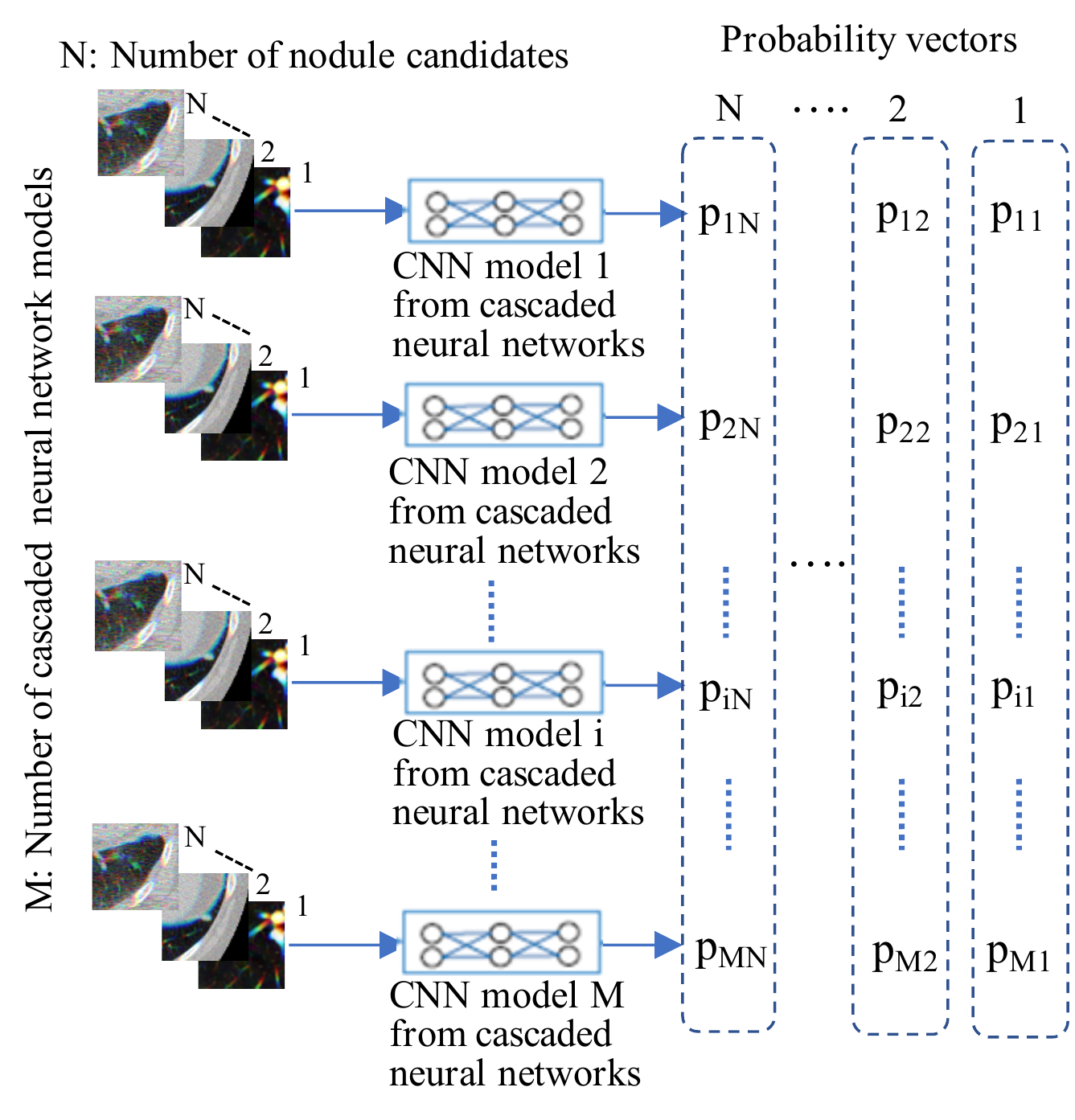}
\caption{Calculation of nodule probability vectors.}
\label{fig:fig2}
\end{figure}

\begin{figure}
\centering
\includegraphics[width=0.5\textwidth]{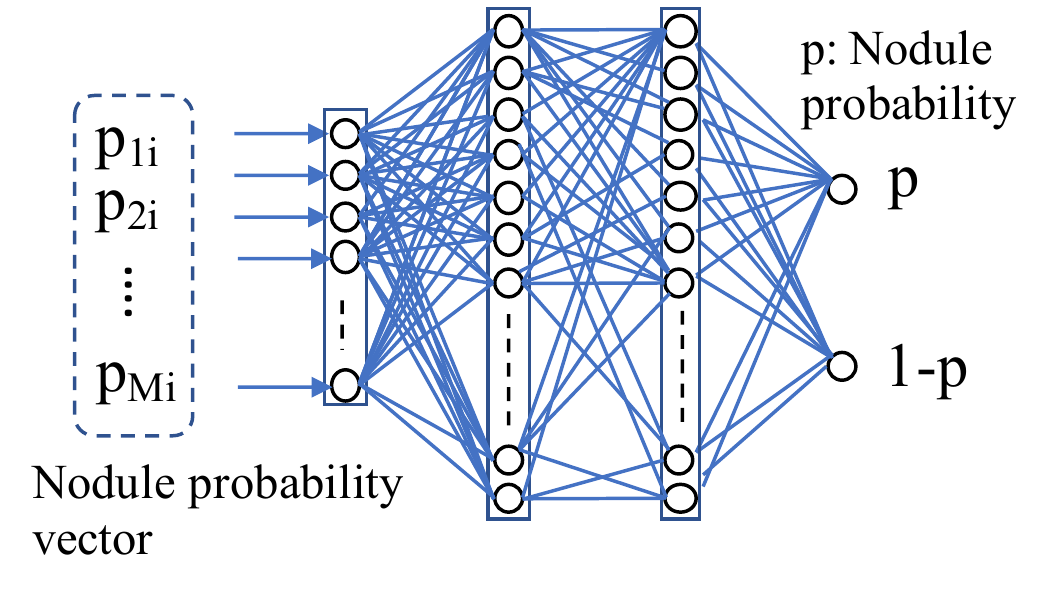}
\caption{Fusion classifier}
\label{fig:fig3}
\end{figure}

Figure \ref{fig:fig2} shows a schematic diagram of probability vector generation. A nodule candidate image is input to M CNN models, thereby a nodule probability vector with length M is obtained. If the total number of nodule candidate images is N, N probability vectors are obtained.
Figure \ref{fig:fig3} shows Fusion classifier with four-layer deep neural network. The intermediate layers are fully connected, but the drop out is 0.5. The number of nodes of second layer and third layer is 70 and 20 respectively and ReLU (Rectified Linear Unit) is used for activity function. In the output later, a softmax is used for activity function. The number of nodes of output layer is 2.

\section{Experiments}
\subsection{Lung CT image data set}
We used the lung CT scan data set obtained from Lung Nodule Analysis 2016 Challenge (LUNA2016) \cite{Grandchalengeluna16}. The data set includes 888 CT scan images along with annotations that were collected during a two-phase annotation process overseen by four experienced radiologists. Each radiologist marked lesions, they identified as non-nodule, nodule < 3 mm, and nodule ≧ 3 mm. The data set consists of all nodules ≧ 3 mm accepted by at least 3 out of 4 radiologists. The complete data set is divided into ten subsets to be used for the 10-fold cross-validation. For convenience, the corresponding class label (0 for non-nodule and 1 for nodule) for each candidate is provided. 1,348 lesions are labeled as nodules and the other 551,065 are non-nodule lesions. In this study, center coordinates of each lesion are given. We use three consecutive slices to obtain volumetric information. Size of each image cropped from CT scan images is 48 pixels × 48 pixels.

\subsection{Experimental setup}
\begin{figure}
\centering
\includegraphics[width=0.6\textwidth]{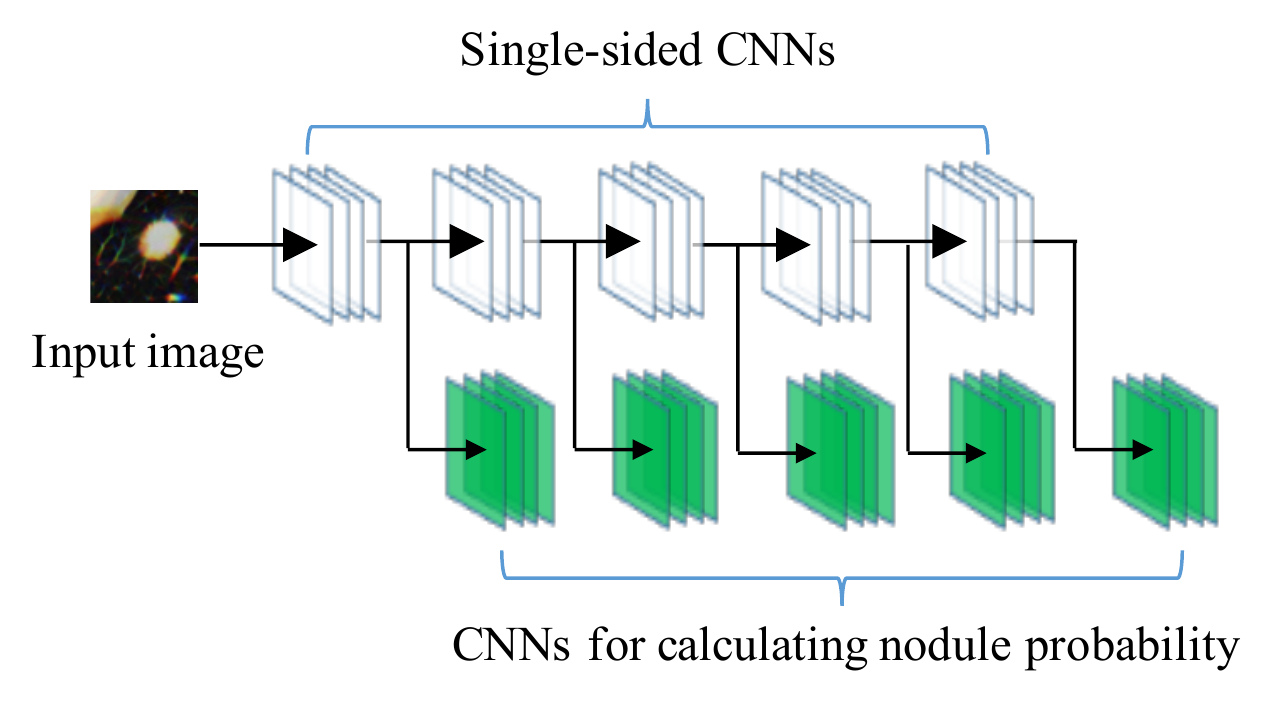}
\caption{Cascaded CNNs which consists of five single-sided classifiers and five balanced CNNs.}
\label{fig:fig4}
\end{figure}

In the cascaded neural networks, CNN models that perform as single-sided classifiers are trained by inversely imbalanced data set as described in section 3. We generated ten inversely imbalanced subsets from LUNA2016 data set. Each subset consists of non-nodule images which are randomly subsampled to 100 samples and nodule images which are oversampled nine times by randomly rotating and scaling for the all nodule images. Thereby, the ratio of the number of nodules and non-nodules was about 12:1. A CNN model at the final stage was trained by a balanced data set. Each cascaded CNN model was trained and tested by using cross-validation manner. In the cross-validation, eight subsets were used for training, one subset was used for calculating accuracy of each cascaded CNN model and the remaining subset was used for testing the cascaded CNN model. We used cascaded CNNs which consists of five single-sided classifiers and five balanced CNNs as shown in Figure  \ref{fig:fig4}. By using these ten CNN models, we generated nodule probability vectors with length M=10. The nodule vectors were calculated for N=551,065 images. Fusion classifier was trained and evaluated by using 10-fold class-validation manner with the nodule probability vectors.

\section{Experimental results}
\begin{figure}
\centering
\includegraphics[width=0.6\textwidth]{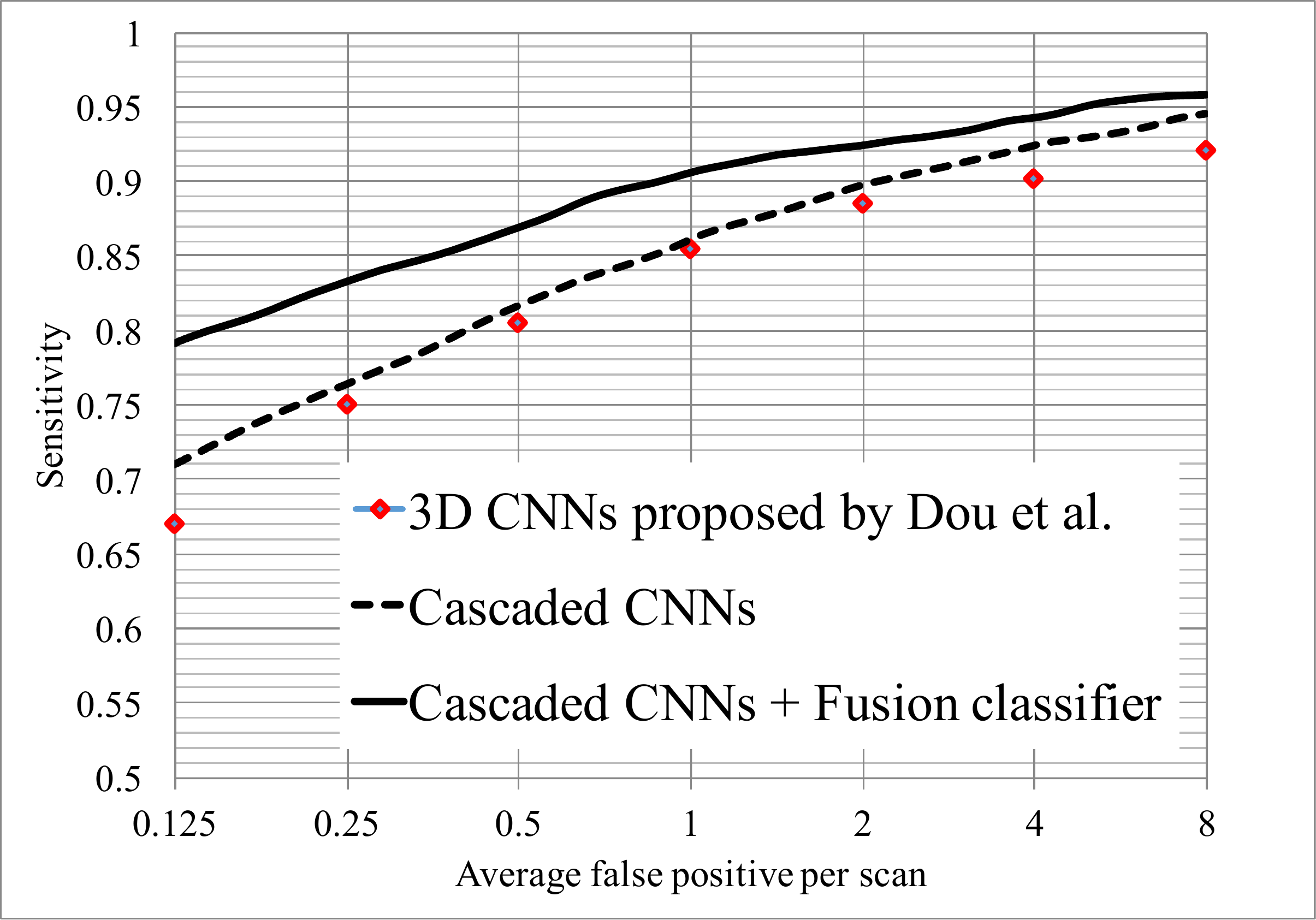}
\caption{FROC curves of proposed method (solid line), cascaded CNNs (dotted line) and Dou‘s method (marks).}
\label{fig:fig5}
\end{figure}

We evaluated classification performance by measuring the sensitivity and average false positive rate per scan by using the free receiver operation characteristic (FROC) analysis provided by LUNA2016. Figure \ref{fig:fig5} shows FROC curves of Fusion classifier in conjunction with the cascaded CNN models (solid line), cascaded CNNs (dotted line) and 3D CNNs proposed by Dou et al. (marks). The performance of the cascade CNNs already exceeded the result of Dou et al., however, the performance is further improved by constructing Fusion classifier using the output of the cascade CNNs. Fusion classifier achieves the sensitivity of 94.4\% and 95.9\% at 4 and 8 false positives per scan, respectively.

\section{Conclusion}
In this paper, we have presented cascaded multi-stage neural networks with single-sided classifiers to reduce the false positives of lung nodule classification in CT scan images. We have shown that the proposed method is better than state of the art CNN methods proposed by Setio, et al. \cite{setio2016pulmonary} and Dou, et al. \cite{dou2016multi}.
Our method can decrease the burden of image interpretation on radiologists. In principle, our method is a kind of boosting method. We believe it can be applied to other class imbalanced problems. 

\end{document}